\documentclass[11pt]{article}

\usepackage[preprint]{acl}

\usepackage{times}
\usepackage{latexsym}
\usepackage{amsmath} 
\usepackage{adjustbox}
\usepackage{makecell}
\usepackage[T1]{fontenc}
\usepackage{bbding}
\usepackage{algpseudocode}
\usepackage[utf8]{inputenc}
\usepackage{booktabs}
\usepackage{multirow}
\usepackage{pifont}
\usepackage{microtype}
\usepackage{listings}
\usepackage{inconsolata}
\usepackage{colortbl}
\usepackage{tcolorbox}
\usepackage{bookmark}

\usepackage{graphicx}
\usepackage[ruled,lined]{algorithm2e}

\usepackage{cleveref}
\title{From LLM-anation to LLM-orchestrator: Coordinating Small Models for Data Labeling}

\author{
Yao Lu\textsuperscript{1,2}\thanks{Work done when intern at A*STAR.}
\quad Zhaiyuan Ji\textsuperscript{1} 
\quad Jiawei Du\textsuperscript{2}
\quad Yu Shanqing\textsuperscript{1}\\
\quad \textbf{Qi Xuan}\textsuperscript{1}\thanks{Corresponding author: Qi Xuan, Tianyi Zhou.}
\quad \textbf{Tianyi Zhou}\textsuperscript{2}
\\
\textsuperscript{1}{Zhejiang University of Technology}
\quad \textsuperscript{2}{Agency for Science, Technology and Research} 
}

\begin{document}
\maketitle

\begin{abstract}

Although the annotation paradigm based on Large Language Models (LLMs) has made significant breakthroughs in recent years, its actual deployment still has two core bottlenecks: first, the cost of calling commercial APIs in large-scale annotation is very expensive; second, in scenarios that require fine-grained semantic understanding, such as sentiment classification and toxicity classification, the annotation accuracy of LLMs is even lower than that of Small Language Models (SLMs) dedicated to this field. To address these problems, we propose a new paradigm of \textbf{multi-model cooperative annotation} and design a fully automatic annotation framework \textbf{\emph{AutoAnnotator}} based on this. Specifically, \emph{AutoAnnotator} consists of two layers. The upper-level meta-controller layer uses the generation and reasoning capabilities of LLMs to select SLMs for annotation, automatically generate annotation code and verify difficult samples; the lower-level task‑specialist layer consists of multiple SLMs that perform annotation through multi-model voting. In addition, we use the difficult samples obtained by the secondary review of the meta-controller layer as the reinforcement learning set and fine-tune the SLMs in stages through a continual learning strategy, thereby improving the generalization of SLMs. Extensive experiments show that \emph{AutoAnnotator} outperforms existing open-source/API LLMs in zero-shot, one-shot, CoT, and majority voting settings. Notably, \emph{AutoAnnotator} reduces the annotation cost by 74.15\% compared to directly annotating with GPT-3.5-turbo, while still improving the accuracy by 6.21\%. Project page: \url{https://github.com/Zhaiyuan-Ji/AutoAnnotator}.



\end{abstract}

\begin{figure}[t]
    \centering
    \includegraphics[width=0.99\linewidth]{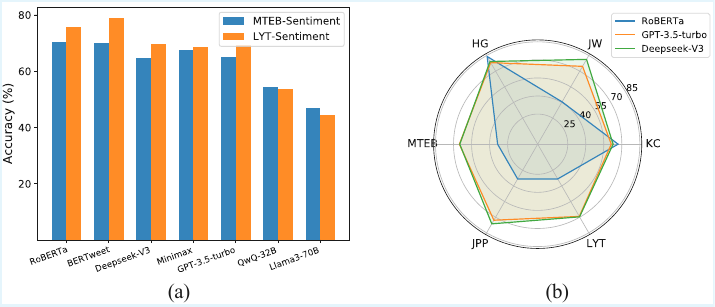}
    \caption{(a) Comparison of classification performance between Large Language Models (LLMs) and Small Language Models (SLMs) on two sentiment classification tasks, proving that SLMs outperform LLMs on domain-related tasks. (b) Classification performance of LLMs and SLMs on $3$ sentiment classification datasets (MTEB-Sentiment, JPP-Sentiment, LYT-Sentiment) and $3$ toxicity classification datasets (KC-Toxicity, JW-Toxicity, HG-Toxicity), illustrating that SLMs exhibit weaker generalization than LLMs.}
    \label{fig:different}
\end{figure}

\begin{figure*}[htbp]
    \centering
    \includegraphics[width=0.99\linewidth]{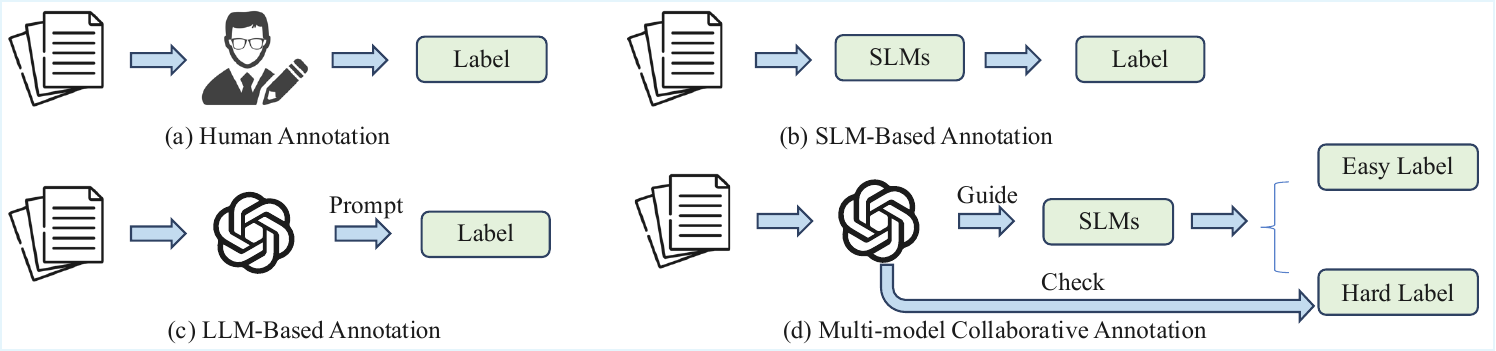}
    \caption{Different data annotation paradigms. (a) represents the traditional manual annotation paradigm. (b) denotes SLM-based annotation paradigm. (c) is the most popular LLM-based annotation paradigm. (d) denotes the multi-model collaborative annotation paradigm proposed by us. Our paradigm can not only improve the annotation accuracy, but also significantly reduce the annotation cost.}
    \label{fig:differentParadigm}
\end{figure*}

\section{Introduction}
High-quality annotated data is key to advancing deep learning~\cite{emam2021state,rasmussen2022challenge,alpaca,ye2025limo}, yet acquiring such data requires specialized domain expertise and is costly~\cite{denton2021whose}, especially when  manually annotating a large number of samples. With the rapid development of LLMs~\cite{achiam2023gpt,guo2025deepseek,team2025kimi}, their powerful semantic understanding~\cite{wu2023next}, contextual reasoning~\cite{sun2024determlr} and generation capabilities~\cite{mo2024large} has driven researchers to develop LLM-based annotation methods~\cite{yadav2024towards,chen2024large,wu2024enhancing,tekumalla2023leveraging,flamholz2024large,laskar2023can,ba2024fill} to reduce the cost of manual annotation.

However, our priori experiments show that this “one-size-fits-all” approach does not work in all areas. In tasks such as sentiment classification~\cite{brauwers2022survey,jiang2011target} and toxicity classification~\cite{van2018challenges,he2024you,li2024hot}, LLMs without special training perform much worse than smaller models that have been specifically fine-tuned (see \Cref{fig:different}(a)). Besides, the annotation cost is often prohibitively expensive, especially when scaling to large datasets. For example, annotating $100,000$ short reviews—each averaging $1024$ input tokens and $20$ output tokens—using GPT-o1 (at $\$15$ per $1$M input tokens and $\$60$ per $1$M output tokens) will cost roughly $\$1,656$. In contrast, the annotation cost of SLMs is almost negligible.


So can we do the opposite: let SLMs (e.g., BERT~\cite{devlin2019bert} and Roberta~\cite{liu2019roberta}) take on the “main force” of data annotation, and efficiently generate initial annotations with its low annotation cost and rich domain knowledge; and when SLMs have low confidence or the sample is more difficult, LLMs will provide secondary review, so as to balance cost-effectiveness and annotation quality. The reason for using LLMs to re-verify difficult examples is that although SLMs outperform LLMs on their familiar domains, their limited generalization ability is unreliable in the case of diverse real data annotations (see \Cref{fig:different}(b)). This \textbf{multi-model collaborative annotation} paradigm can not only reduce overall API overhead, but also leverage the powerful reasoning capabilities of LLMs to improve the labeling accuracy of key samples. \Cref{tab:4paradigms} illustrates the difference between our proposed annotation paradigm and other existing paradigms.

\begin{table}[t]
  \centering
  \caption{Comparison of efficiency, accuracy and generalization ability of different annotation paradigms.}
    \resizebox{0.49\textwidth}{!}{
    \begin{tabular}{ccccc}
    \toprule
          & \multicolumn{1}{c}{Human} & \multicolumn{1}{c}{SLMs} & \multicolumn{1}{c}{LLMs} & \multicolumn{1}{c}{\emph{AutoAnnotator}} \\
    \midrule
    No manual labeling required &   \ding{55}    &   \ding{51}     &    \ding{51}    & \ding{51}  \\
    Low annotation cost &   \ding{55}    &   \ding{51}    &  \ding{55}     & \ding{51} \\
    High labeling accuracy &    \ding{51}   &   \ding{51}    &    \ding{55}   &  \ding{51}\\
    Good generalization &   \ding{51}    &   \ding{55}    &   \ding{51}    &  \ding{51}\\
    \bottomrule
    \end{tabular}}
  \label{tab:4paradigms}%
\end{table}%


In this paper, we introduce a self-evolving \textbf{Auto}mated \textbf{D}ata \textbf{A}nnotation framework, dubbed as \textbf{\emph{AutoAnnotator}}, to improve the existing annotation paradigm. \emph{AutoAnnotator} coordinates both the generative generation and reasoning capabilities of LLMs and the fine-grained task determination capabilities of SLMs, achieving adaptive model selection, automatic code generation, multi-model consensus annotation, and model iterative evolution. By leveraging the low-cost efficiency of SLMs and selectively invoking LLMs only when necessary, our framework significantly reduces annotation cost while achieving superior or comparable annotation quality. By leveraging the low-cost, domain-specific nature of SLMs and selectively invoking LLMs only when necessary, \emph{AutoAnnotator} significantly reduces annotation costs while achieving superior annotation performance.

The entire \emph{AutoAnnotator} framework can be viewed as a hybrid expert system and consists of two layers: a meta-controller layer and a task-specialist layer. Specifically, the meta-controller layer, powered by LLMs, is responsible for selecting appropriate SLMs from Hugging Face\footnote{\url{https://huggingface.co}} based on the given annotation task and automatically generating the code required for the entire annotation process. Since SLMs have limited generalization ability on out-of-domain (difficult) samples, and LLMs have stronger generalization due to pre-training on massive and diverse data, meta-controller will call LLMs to perform a second review of these difficult samples, thereby significantly improving the generalization performance of the overall labeling system. 

The task-specialist layer comprises the selected SLMs by the meta-controller layer. Specifically, each input is fed to all SLMs, and the predictions of these SLMs are aggregated through a majority voting consensus mechanism to generate high-confidence labels. Samples that do not reach the consensus threshold are automatically labeled and returned to the meta-controller layer for secondary verification using the LLM. Once the hard‑sample pool reaches a predefined threshold, these expert‑verified examples trigger an iterative fine‑tuning cycle: each SLM is updated on the collected hard samples, and the refined models then rejoin the consensus pool for subsequent annotation. This continual enhancement loop ensures that the specialists progressively improve their generalization. Overall, our contributions can be summarized as follows:

\begin{itemize}
    \item \textbf{A new paradigm of data annotation.} We propose the paradigm of LLMs guidance with SLMs execution, where LLM uses its powerful generation and reasoning capabilities to build the annotation environment and review the annotation results, while SLMs apply their domain-specific knowledge to carry out the actual labeling.
    \item \textbf{A fully automatic annotation framework.} We introduce a two-layer annotation framework \textbf{\emph{AutoAnnotator}}, which fully automates the annotation model selection, code generation, annotation verification, and annotation model iterative update process.
    \item \textbf{Cost Reduction and Improved Performance.} \emph{AutoAnnotator} outperforms existing opened-source LLMs (7B–70B) and API models (including Minimax, Deepseek-V3, Deepseek-R1, GPT-3.5-turbo and GPT-4o), and consistently maintains optimal performance under multiple labeling strategies such as zero-shot, one-shot, CoT, and majority-vote. Besides, \emph{AutoAnnotator} reduces the annotation cost by 74.15\% compared to directly annotating with GPT-3.5-turbo, while still improving the accuracy by 6.21\%.
    
\end{itemize}

\section{Related Work}
\textbf{LLM-Based Data Annotation.}
Thanks to the remarkable capabilities of LLMs across a wide range of tasks, recent research has gained increased interest in using LLMs for data annotation. For instance, Jadhav et al~\cite{jadhav2024limitations} utilize both closed-source and open-source LLMs to annotate a low-resource language Marathi. Chen et al.~\cite{chen2024large} utilize LLMs to generate samples that are consistent with the data distribution of the benchmark dataset for event extraction, thereby alleviating the challenges of data imbalance and scarcity. Similarly, Li et al~\cite{li2024optimizing} use LLMs for high-quality code retrieval query annotation. Choi et al.~\cite{choi2024multi} extend cost-effective LLM-based annotation beyond traditional data annotation tasks to filter out noisy documents from a multi-document summarization dataset. Liu et al~\cite{liu2025automated} leverage LLMs in combination with historically annotated data and expert-constructed codebooks to extrapolate and extend longitudinal network datasets into future periods. Besides, some studies use LLMs to improve the original annotations made by human annotators~\cite{laskar2023can,flamholz2024large}. Although LLM-based data annotation methods have made significant progress, their application still faces two major challenges: on the one hand, the high cost of API calls makes it difficult to achieve large-scale economy; on the other hand, in tasks that require fine-grained semantic understanding (such as sentiment classification~\cite{brauwers2022survey,jiang2011target} and toxicity classification~\cite{van2018challenges,he2024you,li2024hot}), the annotation performance of LLMs is often inferior to that of specially fine-tuned SLMs.

\begin{figure*}[t]
    \centering
    \includegraphics[width=0.99\linewidth]{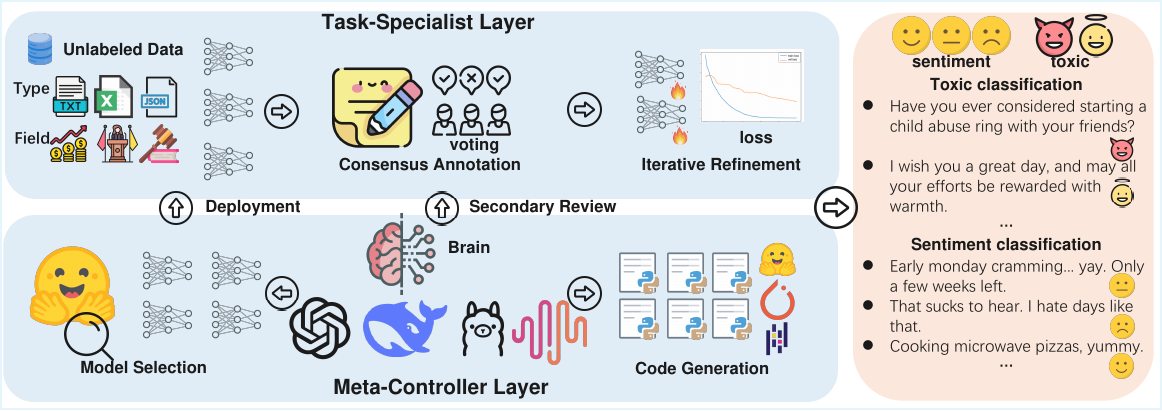}
    \caption{Visualization of the pipeline of \emph{AutoAnnotator}. \emph{AutoAnnotator} consists of two layers: a meta-controller layer and a task-specialist layer. The meta-controller layer, powered by LLMs, is responsible for selecting appropriate SLMs from Hugging Face, automatically generating the code required for the entire annotation process and performing secondary review on samples that are difficult for SLMs. The task-specialist layer comprises the selected SLMs by the meta-controller layer. SLMs use a majority voting mechanism to annotate samples and periodically use difficult samples from LLMs for secondary review to continuously update themselves.}
    \label{fig:pipeline}
\end{figure*}

\noindent\textbf{Collaboration between LLMs and SLMs.}
Collaboration between LLMs and SLMs combines the former’s generalization and reasoning strengths with the latter’s efficient, domain-specific expertise, yielding superior performance and cost-efficiency across various tasks, especially on resource-constrained edge devices. For example, Xu et al.~\cite{xu2023small} use predictions from SLMs to improve LLM in-context learning. CoGenesis~\cite{zhang2024cogenesis} integrates LLMs (hosted on cloud infrastructure) and SLMs (deployed on local devices) to address privacy concerns logically. CITER~\cite{zheng2025citer} adopts a token-level routing strategy, routing non-critical tokens to the SLM to improve effciency, while routing critical tokens to the LLM to ensure generation quality. Collab-RAG~\cite{xu2025collab} employs an SLM to decompose complex queries and improves the SLM's decomposition ability through feedback signals provided by a black-box LLM. Glocker et al.~\cite{glocker2025llm} use a task-specific LLM as the “brain” to drive multiple field-specialized SLMs to perform sub-tasks such as routing and task planning. Inspired by existing studies on LLMs and SLMs collaboration, we innovate the existing LLM-based annotation framework and propose a two-layer automated annotation system, with LLMs as guidance and SLMs as execution.



\section{Method}
\label{sec:method}

In this section, we delve into the \textbf{\emph{AutoAnnotator}}, a hierarchical system that synergizes LLMs with SLMs for automated data annotation. As illustrated in \Cref{fig:pipeline}, the system operates through two interdependent layers: the \textbf{Meta-Controller Layer} and the \textbf{Task-Specialist Layer}. This design complements the powerful generation and reasoning capabilities of the LLMs with the efficient domain expertise of the SLMs, not only achieving better annotation performance, but also significantly reducing annotation costs.

\subsection{Meta-Controller Layer}
The Meta-Controller Layer serves as the decision-making unit that orchestrates the entire annotation process. It mainly implements three core functions: adaptive model selection, automatic code generation and difficult sample verification. Next, we will introduce these functions in detail.

\noindent\textbf{Adaptive Model Selection.} Assuming there is a dataset $\mathcal{D}=\left\{x_1, x_2, \ldots, x_n\right\}$ to be annotated, \emph{AutoAnnotator} first needs to determine SLMs for annotation. However, faced with millions of open-source models\footnote{As of May $12$, $2025$, there are $1,685,478$ open source models on HuggingFace.} on platforms such as HuggingFace\footnote{https://huggingface.co/}, non-professionals often find it difficult to filter out models that are suitable for the current task from the complex model descriptions. To address this challenge, we built an adaptive model selection engine using LLMs, eliminating human intervention in model selection. Specifically, given an annotation task $T$, we utilize the LLM to give a list of task-related model recommendations by querying about $1.69$M HuggingFace models, and take the Top-$k$ models for annotation. This process can be formulated as follows:
\begin{equation}
\mathcal{M}_t=\operatorname{Top-k}\left(\operatorname{sim}\left(f_{\operatorname{LLM}}\left(T\right), f_{\operatorname{LLM}}(d)\right)\right),
\end{equation}
where $d$ denotes the description of the corresponding model.

\noindent\textbf{Automatic Code Generation.} After obtaining the recommendation list, an intuitive method is to download and deploy these models locally. Then, we can start the data annotation and subsequent processing steps. However, in this workflow, many processes usually require manual programming, such as SLMs deployment, data annotation, and SLMs fine-tuning, which makes the entire process labor-intensive. To address this limitation and maximize automation, we equip the meta-controller layer with an automatic code generation capability. Given the powerful code generation capabilities of LLMs~\cite{wang2023review,roziere2023code,jiang2024survey}, we directly prompt it to generate all the scripts required for the annotation pipeline.

\noindent\textbf{Difficult Sample Verification.} While SLMs exhibit superior performance on domain-specific annotation tasks, they often struggle with out-of-domain samples. In other words, SLMs have limited generalization ability, making them less reliable when performing complex annotation tasks. In contrast, LLMs, especially those like GPTs and DeepSeek, trained on diverse data, show stronger generalization capabilities~\cite{li2025zero} and can better handle more complex situations (see ). To this end, \emph{AutoAnnotator} leverages LLMs in the meta-controller layer to perform secondary validation on complex or uncertain samples, achieving both high accuracy and strong generalization across various data conditions.

\subsection{Task-Specialist Layer}
The Task-Specialist Layer is responsible for the actual annotation, using a set of lightweight, domain-specific pre-trained SLMs to efficiently label the data. It consists of two components: the Multi-Model Consensus Annotation module and the Expert-Guided Iterative Refinement module.

\noindent\textbf{Multi-Model Consensus Annotation.}
As mentioned earlier, SLMs exhibit poor generalization ability. To address this, we aggregate predictions from a diverse pool of SLMs and only accept labels on which they reach high agreement. Formally, let $\mathcal{D}=\{x_i\}_{i=1}^{N}$ be the unlabeled dataset and $\mathcal{M}=\{m_1,m_2,\cdots,m_k\}$ be the pool of SLMs recommended from the Meta-Controller Layer. For each data sample $x_i \in \mathcal{D}$, run all $k$ models in parallel to obtain their annotation results:
\begin{equation}
\mathcal{Y}_i=\left\{y_i^{(1)}, y_i^{(2)}, \ldots, y_i^{\left(k\right)}\right\}.
\end{equation}
The final label $\hat{y}_i$ is determined by majority voting of these $k$ models:
\begin{equation}
\hat{y}_i=\operatorname{Majority Voting}\left(\mathcal{Y}_i\right).
\end{equation}
In order to evaluate the degree of consensus among models, the uncertainty metric $\mathcal{U}$ is introduced, which is defined as:
\begin{equation}
\mathcal{U}\left(x_i\right)=1-\frac{\max _y \#\left\{y_i^{(j)}=y\right\}}{k},
\end{equation}
where $\#\left\{y_i^{(j)}=y\right\}$ represents the number of models that predict $y$. If $\mathcal{U}\left(x_i\right)$ is greater than the predefined value $\epsilon$, the sample is considered to have a large disagreement and is automatically stored in the secondary review pool $\mathcal{D}_{\text{hard}}$:
\begin{equation}
\operatorname{Route}(x)= \begin{cases}\operatorname{Direct \, Labeling} & \mathcal{U}(x)<\epsilon \\ \operatorname{Secondary \, Review} & \mathcal{U}(x) \geq \epsilon.\end{cases}
\end{equation}
The Task-Specialist Layer dynamically selects between two verification modes (automatic LLM annotation or manual human annotation) based on user needs and cost-accuracy trade-off analysis.

\noindent\textbf{Expert-Guided Iterative Refinement.}
To explicitly enhance the generalization of specialist SLMs on difficult (out‑of‑domain) samples, we introduce a continual fine‑tuning procedure guided by expert labels (from either LLMs or human annotators). Specifically, once the number of samples in the hard sample pool $\mathcal{D}_{\text{hard}}$ reaches a predefined size $\beta$, we pause the consensus annotation and start the continual fine-tuning cycle. 
For each SLM $m_i \in \mathcal{M}$, we fine‑tune it on $\mathcal{D}_{\text{hard}}$ for up to a budgeted number of epochs, producing updated specialists $m'_i$. Taking the model $m_i$ as an example, we define the loss function as follows:
\begin{equation}
\mathcal{L}\left(\theta\right)=\sum\operatorname{CE}(m_i(x;\theta), y), \quad {(x, y)\in\mathcal{D}_{\text{hard}}}
\end{equation}
where $\operatorname{CE}(\cdot)$ represents the cross-entropy loss. Then gradient descent is used to update the parameters:
\begin{equation}
\theta \leftarrow \theta-\alpha \nabla_\theta \mathcal{L}\left(\theta\right),
\end{equation}
where $\alpha$ denotes the learning rate. Subsequently, the refined model $m'_i$ replaces $m_i$ for annotation. After that, we resume annotation on remaining unlabeled data until $\mathcal{D}_{\text{hard}}$ exceeds $\beta$ again, triggering a further cycle of continuous fine-tuning.

\begin{table*}[t]
  \centering
  \caption{The appropriate Hugging Face models selected by LLMs based on the type of annotation task.}
    \resizebox{0.85\textwidth}{!}{
    \begin{tabular}{c|c|c|c|c}
    \toprule
    \multicolumn{1}{c|}{Task Type} & Model ID & Parameters & HF Downloads & Nick name\\
    \midrule
    \multirow{3}[2]{*}{Sentiment} & cardiffnlp/twitter-roberta-base-sentiment-latest & 125M  & 2.43M &SLM1\\
          & cardiffnlp/twitter-xlm-roberta-base-sentiment & 125M  & 2.06M &SML2\\
          & finiteautomata/bertweet-base-sentiment-analysis & 110M  & 1.06M &SML3\\
    \midrule
    \multirow{3}[2]{*}{Toxicity} & s-nlp/roberta\_toxicity\_classifier & 110M  & 160K &SLM1\\
          & JungleLee/bert-toxic-comment-classification & 110M  & 46.3K &SLM2\\
          & garak-llm/toxic-comment-model & 67M  & 9.67K &SLM3\\
    \bottomrule
    \end{tabular}}
  \label{tab:info}%
\end{table*}%

\section{Experiments}
\subsection{Experimental Settings}
\textbf{Datasets.} To evaluate the effectiveness of \emph{AutoAnnotator}, we conduct extensive experiments on two representative annotation tasks, sentiment classification and toxicity classification, using a total of six real‑world datasets. Specifically, in this study, we select mteb/tweet\_sentiment\_extraction, jppgks/twitter-financial-news-sentiment and LYTinn/sentiment-analysis-tweet for the sentiment classification task and karthikarunr/Cyberbullying-Toxicity-Tweets, jiaxin-wen/Implicit-Toxicity and heegyu/toxic\_conversations\_balanced for the toxicity classification task. We provide a detailed introduction to these dataset in the Appendix~\ref{sec:Dataset_Description}. To simplify the description, we use the following aliases for these datasets: MTEB-Sentiment, JPP-Sentiment, LYT-Sentiment, KC-Toxicity, JW-Toxicity, and HG-Toxicity, respectively.

\noindent\textbf{Models.} In \emph{AutoAnnotator}, all SLMs involved in annotation are automatically selected by the LLM in the meta-controller layer based on task characteristics. In this paper, we select $3$ SLMs for each annotation task by default. We provide details of the model selected by the LLM in \Cref{tab:info}. 

\noindent\textbf{Implementation Details.} By default, we use $3$ ($k=3$) SLMs for each annotation task. Once the hard-sample pool $\mathcal{D}_{\text{hard}}$ reaches a predefined threshold $\beta=2,000$, we will pause to fine‑tune all $3$ specialists on $\mathcal{D}_{\text{hard}}$, ensuring they can continually learn new things throughout the annotation process. 

We provide detailed ablation experiments in \Cref{sec:Ablation Study}. As for fine-tuning the SLMs, we set the initial learning rate, weight decay and epoch to $2e-5$, $0.01$ and $3$, respectively. All experiments are conducted on 2 NVIDIA A100.

\noindent\textbf{Methods for Comparison.} To evaluate the effectiveness of \emph{AutoAnnotator}, we compare it with three types of baselines: (1) SLMs selected by the LLM in the meta-controller layer (2) Open-source LLMs, range from Mistral-7B, Baichuan2-7B~\cite{yang2023baichuan}, Qwen2.5-7B-Instruct~\cite{yang2024qwen2}, Llama3.1-8B-Instruct~\cite{grattafiori2024llama}, Llama2-13B~\cite{touvron2023llama}, QwQ-32B and Llama3-70B~\cite{grattafiori2024llama}. (3) API models, such as MiniMax, DeepSeek V3~\cite{liu2024deepseek}, DeepSeek R1~\cite{guo2025deepseek} and GPT-3.5-Turbo. We provide all prompts used by baselines and \emph{AutoAnnotator} in \Cref{table:apiprompt} and \Cref{table:sysprompt}.

\begin{table*}[t]
  \centering
  \caption{Comparison of the proposed \emph{AutoAnnotator} with existing methods on different toxicity and sentiment annotation tasks. It is worth noting that the SLM1 for sentiment classification and the SLM1 for toxicity classification are not the same model. The specific models of each model are shown in \Cref{tab:info}.}
  \resizebox{1\textwidth}{!}{
    \begin{tabular}{ccccccccccc}
    \toprule
          & \multicolumn{5}{c}{Sentiment Classification} & \multicolumn{5}{c}{Toxicity Classification} \\
    \midrule
    Model & MTEB-Sentiment & JPP-Sentiment &  LYT-Sentiment & Avg   & \# LLM Calls &  KC-Toxicity & JW-Toxicity & HG-Toxicity & Avg   & \# LLM Calls \\
    \midrule
    \rowcolor[rgb]{ .929,  .929,  .929} \multicolumn{11}{c}{SLMs Only} \\
    \midrule
    SLM1  & 70.43\% & 70.31\% & 75.54\% & 72.09\% & 0     & 66.88\% & 40.56\% & 84.06\% & 63.83\% & 0 \\
    SLM2  & 69.55\% & 59.76\% & 72.52\% & 67.28\% & 0     & 59.27\% & 44.07\% & 80.87\% & 61.40\% & 0 \\
    SLM3  & 70.17\% & 69.22\% & 78.83\% & 72.74\% & 0     & 68.13\% & 39.30\% & 67.07\% & 58.17\% & 0 \\
    \midrule
    \rowcolor[rgb]{ .929,  .929,  .929} \multicolumn{11}{c}{Open-source LLMs, Zero-shot} \\
    \midrule
    Mistral-7B-V0.2 & 35.91\% & 23.24\% & 28.37\% & 29.17\% & 38396 & 68.43\% & 56.32\% & 55.14\% & 59.96\% & 48475 \\
    Baichuan2-7B-Base & 38.66\% & 24.96\% & 30.38\% & 31.33\% & 38396 & 20.99\% & 61.22\% & 67.68\% & 49.96\% & 48475 \\
    Qwen2.5-7B-Instruct & 65.62\% & 74.62\% & 67.66\% & 69.30\% & 38396 & 63.92\% & 76.25\% & 77.20\% & 72.46\% & 48475 \\
    Llama3.1-8B-Instruct & 51.90\% & 57.37\% & 53.15\% & 54.14\% & 38396 & 72.08\% & 63.14\% & 60.03\% & 65.08\% & 48475 \\
    Llama2-13B & 48.85\% & 38.90\% & 51.02\% & 46.26\% & 38396 & 55.29\% & 75.71\% & 67.08\% & 66.03\% & 48475 \\
    QwQ-32B & 54.28\% & 69.01\% & 53.72\% & 59.00\% & 38396 & 70.81\% & 77.87\% & 72.61\% & 73.76\% & 48475 \\
    Llama3-70B & 46.82\% & 39.87\% & 44.33\% & 43.67\% & 38396 & 49.62\% & 70.60\% & 69.56\% & 63.26\% & 48475 \\
    \midrule
    \rowcolor[rgb]{ .929,  .929,  .929} \multicolumn{11}{c}{Open-source LLMs, One-shot} \\
    \midrule
    Mistral-7B-V0.2 & 45.24\% & 40.58\% & 44.57\% & 43.46\% & 38396 & 78.73\% & 69.86\% & 55.70\% & 68.10\% & 48475 \\
    Baichuan2-7B-Base & 40.30\% & 66.46\% & 33.37\% & 46.71\% & 38396 & 77.60\% & 87.40\% & 58.88\% & 74.63\% & 48475 \\
    Qwen2.5-7B-Instruct & 67.03\% & 73.62\% & 65.32\% & 68.66\% & 38396 & 58.13\% & 79.18\% & 71.06\% & 69.46\% & 48475 \\
    Llama3.1-8B-Instruct & 59.30\% & 63.36\% & 58.37\% & 60.34\% & 38396 & 58.65\% & 73.83\% & 64.25\% & 65.58\% & 48475 \\
    Llama2-13B & 58.50\% & 68.17\% & 63.04\% & 63.24\% & 38396 & 66.98\% & 88.79\% & 74.11\% & 76.63\% & 48475 \\
    QwQ-32B & 55.59\% & 68.47\% & 37.16\% & 53.74\% & 38396 & 62.50\% & 87.80\% & 76.19\% & 75.50\% & 48475 \\
    Llama3-70B & 49.58\% & 58.08\% & 50.93\% & 52.86\% & 38396 & 65.05\% & 71.02\% & 64.58\% & 66.88\% & 48475 \\
    \midrule
    \rowcolor[rgb]{ .929,  .929,  .929} \multicolumn{11}{c}{Open-source LLMs, CoT} \\
    \midrule
    Mistral-7B-V0.2 & 40.46\% & 23.91\% & 34.20\% & 32.86\% & 38396 & 49.74\% & 74.90\% & 69.29\% & 64.64\% & 48475 \\
    Baichuan2-7B-Base & 36.53\% & 23.74\% & 29.49\% & 29.92\% & 38396 & 47.55\% & 80.47\% & 66.70\% & 64.91\% & 48475 \\
    Qwen2.5-7B-Instruct & 55.44\% & 68.72\% & 63.52\% & 62.56\% & 38396 & 60.27\% & 68.12\% & 76.19\% & 68.19\% & 48475 \\
    Llama3.1-8B-Instruct & 51.74\% & 58.17\% & 53.75\% & 54.55\% & 38396 & 47.93\% & 62.54\% & 64.56\% & 58.34\% & 48475 \\
    Llama2-13B & 51.24\% & 36.22\% & 45.93\% & 44.46\% & 38396 & 57.74\% & 71.31\% & 69.99\% & 66.35\% & 48475 \\
    QwQ-32B & 50.15\% & 68.43\% & 55.20\% & 57.93\% & 38396 & 67.79\% & 80.53\% & 75.68\% & 74.67\% & 48475 \\
    Llama3-70B & 46.81\% & 42.71\% & 46.94\% & 45.49\% & 38396 & 50.61\% & 69.66\% & 71.07\% & 63.78\% & 48475 \\
    \midrule
    \rowcolor[rgb]{ .929,  .929,  .929} \multicolumn{11}{c}{API Models} \\
    \midrule
    Deepseek-V3 & 64.82\% & 76.38\% & 69.68\% & 70.29\% & 38396 & 62.74\% & 81.30\% & 79.27\% & 74.44\% & 48475 \\
    Deepseek-R1 & 66.57\% & 74.25\% & 67.90\% & 69.57\% & 38396 & 75.31\% & 74.26\% & 77.10\% & 75.56\% & 48475 \\
    Minimax-abab6.5s-chat & 67.37\% & 77.22\% & 68.58\% & 71.06\% & 38396 & 72.37\% & 68.18\% & 76.14\% & 72.23\% & 48475 \\
    GPT-3.5-turbo & 65.14\% & 72.91\% & 69.26\% & 69.10\% & 38396 & 61.20\% & 74.63\% & 78.21\% & 71.35\% & 48475 \\
    \midrule
    \rowcolor[rgb]{ .929,  .929,  .929} \multicolumn{11}{c}{LLMs,  Majority Vote} \\
    \midrule
    7 Open-source LLMs Voting (zero-shot) & 54.69\% & 55.86\% & 57.65\% & 56.07\% & 268772 & 65.20\% & 76.63\% & 77.09\% & 72.97\% & 339325 \\
    7 Open-source LLMs Voting (one-shot) & 63.20\% & 71.23\% & 59.58\% & 64.67\% & 268772 & 69.72\% & 86.73\% & 72.99\% & 76.48\% & 339325 \\
    4 API Models Voting & 67.78\% & 77.85\% & 70.80\% & 72.14\% & 153584 & 72.07\% & 72.09\% & 79.47\% & 74.54\% & 193900 \\
    \midrule
    \rowcolor[rgb]{ .929,  .929,  .929} \multicolumn{11}{c}{AutoAnnotator (Ours)} \\
    \midrule
    AutoAnnotator+Minimax & 67.78\% & 81.20\% & 74.80\% & 74.59\% & 10643 & 73.73\% & 73.75\% & 83.55\% & 77.01\% & 18210 \\
    AutoAnnotator+Deepseek-V3 & 66.77\% & 77.96\% & 73.61\% & 72.78\% & 10537 & 61.29\% & 85.29\% & 83.25\% & 76.61\% & 17886 \\
    AutoAnnotator+GPT-3.5-turbo & 67.89\% & 78.56\% & 72.90\% & 73.12\% & 10065 & 67.69\% & 82.02\% & 82.97\% & 77.56\% & 18942 \\
    AutoAnnotator+Human & 78.33\% & 82.83\% & 84.13\% & 81.76\% & 8     & 83.50\% & 89.66\% & 91.78\% & 88.31\% & 8 \\
    \bottomrule
    \end{tabular}}
  \label{tab:SOTA}
\end{table*}%

\subsection{Main Experiment Results}
\textbf{Comparison with SLMs.} As shown in \Cref{tab:SOTA}, under the “SLMs Only” setting, the strongest SLM achieves $72.74\%$ average accuracy on sentiment tasks and $63.83\%$ on toxicity. By integrating these SLMs into our \emph{AutoAnnotator} framework, we boost sentiment accuracy to $74.59\%$ ($+1.85\%$) and toxicity to $77.56\%$ ($+13.73\%$). These experimental results demonstrate that \emph{AutoAnnotator} can significantly improve the performance of SLMs on sentiment classification and toxicity classification annotation tasks.

\begin{table*}[t]
  \centering
  \caption{Comparison of annotation cost and efficiency between API models and \emph{AutoAnnotator}.}
  \resizebox{0.95\textwidth}{!}{
    \begin{tabular}{ccccccc}
    \toprule
    Model & Token (Input+Output) & GPU Memory & Time Cost & Time Reduction & Cost & Cost Reduction \\
    \midrule
    Deepseek-V3 & 88023  & - & 71.19 minutes &   -    & 0.027202 \$  & - \\
    \emph{AutoAnnotator}+Deepseek-V3 & 25629  & 4458 MB  & 26.06 minutes  & 63.40\%  & 0.008085 \$  & 70.28\%  \\
    \midrule
    Deepseek-R1 & 356532  & - & 212.93 minutes &   -    & 0.650334 \$ & - \\
    \emph{AutoAnnotator}+Deepseek-R1 & 109666  & 4458 MB  & 39.40 minutes & 81.50\%  & 0.211581 \$ & 67.47\%  \\
    \midrule
    Minimax & 91724  & - & 93.70 minutes &    -   & 0.012723 \$ &  -\\
    \emph{AutoAnnotator}+Minimax & 19357  & 4458 MB & 30.67 minutes & 67.27\%  & 0.003112 \$ & 75.54\%  \\
    \midrule
    GPT-3.5-turbo & 88514  & - & 34.17 minutes &   -    & 0.048423 \$ &  -\\
    \emph{AutoAnnotator}+GPT-3.5-turbo & 22235  & 4458 MB & 17.75 minutes & 48.05\%  & 0.012519 \$ & 74.15\%  \\
    \midrule
    GPT-4 & 88027  & - & 30.56 minutes  &    -   & 2.751180 \$ & - \\
    \emph{AutoAnnotator}+GPT-4 & 23843  & 4458 MB & 18.34 minutes & 40.00\%  & 0.754470 \$ & 72.58\%  \\
    \midrule
    GPT-4o & 87915  & - & 23.20 minutes &   -    & 0.246540 \$ & - \\
    \emph{AutoAnnotator}+GPT-4o & 22087  & 4458 MB & 15.21 minutes & 34.44\%  & 0.064300 \$  & 73.92\%  \\
    \bottomrule
    \end{tabular}}
  \label{tab:cost_comparation}%
\end{table*}%

\begin{figure*}[htbp]
    \centering
    \includegraphics[width=0.9\linewidth]{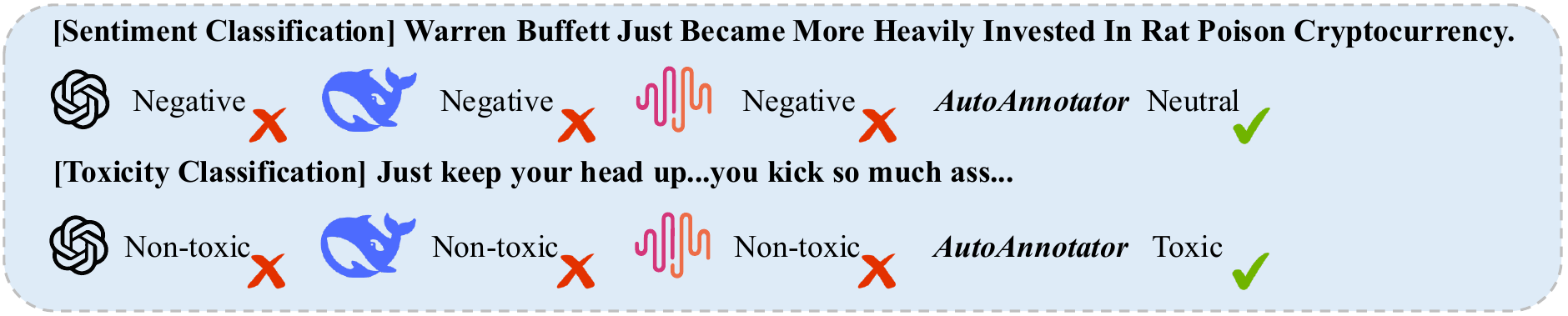}
    \caption{Visualization of representative samples correctly annotated by \emph{AutoAnnotator} but misclassified by APIs.}
    \label{fig:example}
    \vspace{-5mm}
\end{figure*}

\noindent\textbf{Comparison with Open-sourced LLMs.} To benchmark \emph{AutoAnnotator} against open‑source LLMs, we evaluate the latter in three widely used annotation settings, zero‑shot, one‑shot, and chain‑of‑thought~\cite{wei2022chain} (CoT) prompting, on both sentiment and toxicity classification tasks. As for the one‑shot setting, each model is given a single in‑context example before annotation. As for the CoT setting, we add a CoT prompt “Let's think step by step like an operations research expert.” behind the zero-shot prompt. Among these settings, we find that one‑shot prompting consistently outperforms zero‑shot, as the single in‑context example helps the model calibrate its label distributions and reduces misunderstanding of the task. By contrast, chain‑of‑thought prompting hints only marginally improve annotation accuracy, which we believe is because the generated step-by-step reasoning shifts the model's focus away from the classification task. Overall, \emph{AutoAnnotator} consistently outperforms zero‑shot, one‑shot, and chain‑of‑thought prompting strategies, demonstrating its superior annotation accuracy and validating the effectiveness of our multi‑model collaborative paradigm.

\noindent\textbf{Comparison with API Models.}
We further benchmark against API models, including Minimax, Deepseek‑V3, Deepseek-R1 and GPT‑3.5‑turbo. We report our main results in \Cref{tab:SOTA}. The strongest Minimax achieves $71.06\%$ average accuracy on sentiment tasks, while Deepseek‑V3 leads toxicity at $74.44\%$. By integrating these API models into \emph{AutoAnnotator}, we boost sentiment accuracy to $74.59\%$ ($+3.53\%$) and toxicity to $76.61\%$ ($+2.17\%$). Similarly, when GPT-3.5-turbo is used alone, the sentiment accuracy reaches $69.10\%$ and the toxicity accuracy reaches $71.35\%$; when integrated into \emph{AutoAnnotator}, the sentiment accuracy rises to $73.12\%$ ($+4.02\%$) and the toxicity accuracy rises to $77.56\%$ ($+6.21\%$). Besides, compared with direct LLM annotation, \emph{AutoAnnotator} significantly reduces the number of LLM calls ($60\%+$ for sentiment tasks and $70\%+$ for toxicity tasks). It is worth noting that \emph{AutoAnnotator} not only outperforms existing API models in terms of performance, but also far exceeds them in terms of annotation cost and efficiency (see below). We provide some samples that \emph{AutoAnnotator} can annotate correctly, but API models annotate incorrectly in \Cref{fig:example}.

\noindent\textbf{Comparison with LLM Majority‑Vote.} We additionally evaluate majority voting across multiple LLMs. As shown in \Cref{tab:SOTA}, \emph{AutoAnnotator}+Minimax needs only $10,643$ API calls to achieve 74.59\% sentiment accuracy, outperforming both open‑source and API‑voting baselines while reducing LLM calls by over 93\%. Regardless of whether we ensemble multiple open‑source or API LLMs—even with zero‑shot or one‑shot voting—\emph{AutoAnnotator} consistently outperforms all voting schemes while using far fewer LLM calls.

\noindent\textbf{The efficiency and cost of \emph{AutoAnnotator}.} To evaluate the annotation cost and efficiency of API models and \emph{AutoAnnotator}, we conduct a quantitative analysis from three dimensions: computing resource consumption (the number of tokens and GPU memory usage), annotation time cost, and economic cost. All experiments are performed on NVIDIA A100 GPUs, and the annotation task scale is uniformly set to $1000$ samples. We conduct experiments on Deepseek-V3, Deepseek-R1, Minimax, GPT-3.5-turbo, GPT-4 and GPT-4o, respectively. As shown in \Cref{tab:cost_comparation}, \emph{AutoAnnotator} reduces the annotation time by $34.44\%$ (GPT-4o) to $81.50\%$ (Deepseek-R1), with an average reduction of $55.85\%$. Besides, the annotation cost is reduced by $75.54\%$ (Minimax) at the highest and $67.47\%$ (Deepseek-R1) at the lowest, with an average saving of $72.32\%$. In general, \emph{AutoAnnotator} has achieved a significant improvement in annotation efficiency and a significant reduction in annotation costs while maintaining annotation quality through the LLMs and SLMs collaborative annotation paradigm.

\begin{table}[t]
  \centering
  \caption{The impact of the number of SLMs used for annotation on the annotation performance.}
    \begin{tabular}{ccccc}
    \toprule
    k     & 2     & 3     & 4     & 5 \\
    \midrule
    Acc   & 73.32\% & 78.56\% & 76.26\% & 76.26\% \\
    \bottomrule
    \end{tabular}%
  \label{tab:k_aba}%
\end{table}%

\subsection{Ablation Study}
\label{sec:Ablation Study}
\noindent\textbf{The number of SLMs used in the task-specialist layer.} To explore the impact of the number of SLMs on the annotation performance, we perform ablations on the JPP-Sentiment dataset using GPT-3.5-turbo as the meta-controller LLM. We vary the number of SLMs $k$ participating in the multi-model consensus annotation from $2$ to $5$ and report the final annotation accuracy in \Cref{tab:k_aba}. We find that the annotation performance is best when $k=3$. Therefore, considering the computational cost and annotation accuracy, we use $k=3$ as the default in this paper. 

\noindent\textbf{The number of samples in the hard sample pool $\mathcal{D}_{\text{hard}}$.} To explore the impact of the number of hard samples used for continuous fine-tuning at each stage on the annotation performance, we perform ablations on the JPP-Sentiment dataset using GPT-3.5-turbo. We set the sample size $\beta$ to $\{500,1000,2000,3000\}$. As shown in \Cref{tab:beta_aba}, we find that the annotation performance peaks at $\beta=2000$, therefore, we adopt $\beta=2000$ as the default hard‑sample batch size in this paper.


\begin{table}[t]
  \centering
  \caption{The impact of the number of hard samples used for continuous fine-tuning at each stage on the annotation performance.}
    \begin{tabular}{ccccc}
    \toprule
    $\beta$    & 500   & 1000  & 2000  & 3000 \\
    \midrule
    Acc   & 76.17\% & 73.91\% & 78.56\% & 73.12\% \\
    \bottomrule
    \end{tabular}%
  \label{tab:beta_aba}%
\end{table}%


\section{Conclusion} 
In this paper, we propose a new paradigm for multi-model collaborative annotation and designs a fully automatic annotation framework \emph{AutoAnnotator} based on it. Specifically, \emph{AutoAnnotator} consists of a meta-controller layer and a task‑specialist layer. Specifically, the meta-controller layer is responsible for recommending appropriate annotation SLMs, generating the code required for annotation, and rechecking difficult samples that cannot be determined by SLMs. while the task‑specialist layer is responsible for the actual annotation. To enhance the generalization of the SLMs, we use the difficult samples obtained from the second verification of the LLM as a reinforcement learning set, and periodically send it to the SLMs for continuous fine-tuning. Extensive experiments demonstrate the effectiveness of \emph{AutoAnnotator} on six datasets.

\noindent\textbf{Limitations.} While promising, there are still some drawbacks of \emph{AutoAnnotator}. The model selection, code generation, and difficult sample review of the entire framework are all driven by LLMs. Therefore, the performance of \emph{AutoAnnotator} depends to a certain extent on the quality of LLMs.


\noindent\textbf{Ethical Statement.} We are aware of the potential ethical concerns of using LLMs as potential labelers in the data annotation process in terms of the perpetuation of existing biases in LLMs. Since LLMs are trained on vast amounts of texts on the Internet, they can unavoidably incorporate the biases present in these data sources. Besides, \emph{AutoAnnotator} involves the usage of OpenAI APIs. We follow the data usage guidelines for interactions with Microsoft Azure’s OpenAI API service. We do not foresee other ethics issues.

\bibliography{custom}

\begin{thebibliography}{44}
\providecommand{\natexlab}[1]{#1}

\bibitem[{Achiam et~al.(2023)Achiam, Adler, Agarwal, Ahmad, Akkaya, Aleman, Almeida, Altenschmidt, Altman, Anadkat et~al.}]{achiam2023gpt}
Josh Achiam, Steven Adler, Sandhini Agarwal, Lama Ahmad, Ilge Akkaya, Florencia~Leoni Aleman, Diogo Almeida, Janko Altenschmidt, Sam Altman, Shyamal Anadkat, et~al. 2023.
\newblock Gpt-4 technical report.
\newblock \emph{arXiv preprint arXiv:2303.08774}.

\bibitem[{Ba et~al.(2024)Ba, Mancenido, and Pan}]{ba2024fill}
Yang Ba, Michelle~V Mancenido, and Rong Pan. 2024.
\newblock Fill in the gaps: Model calibration and generalization with synthetic data.
\newblock \emph{arXiv preprint arXiv:2410.10864}.

\bibitem[{Brauwers and Frasincar(2022)}]{brauwers2022survey}
Gianni Brauwers and Flavius Frasincar. 2022.
\newblock A survey on aspect-based sentiment classification.
\newblock \emph{ACM Computing Surveys}, 55(4):1--37.

\bibitem[{Chen et~al.(2024)Chen, Qin, Jiang, and Choi}]{chen2024large}
Ruirui Chen, Chengwei Qin, Weifeng Jiang, and Dongkyu Choi. 2024.
\newblock Is a large language model a good annotator for event extraction?
\newblock In \emph{Proceedings of the AAAI Conference on Artificial Intelligence}, volume~38, pages 17772--17780.

\bibitem[{Choi et~al.(2024)Choi, Yun, Jin, and Kim}]{choi2024multi}
Juhwan Choi, Jungmin Yun, Kyohoon Jin, and YoungBin Kim. 2024.
\newblock Multi-news+: Cost-efficient dataset cleansing via llm-based data annotation.
\newblock \emph{arXiv preprint arXiv:2404.09682}.

\bibitem[{Denton et~al.(2021)Denton, D{\'\i}az, Kivlichan, Prabhakaran, and Rosen}]{denton2021whose}
Remi Denton, Mark D{\'\i}az, Ian Kivlichan, Vinodkumar Prabhakaran, and Rachel Rosen. 2021.
\newblock Whose ground truth? accounting for individual and collective identities underlying dataset annotation.
\newblock \emph{arXiv preprint arXiv:2112.04554}.

\bibitem[{Devlin et~al.(2019)Devlin, Chang, Lee, and Toutanova}]{devlin2019bert}
Jacob Devlin, Ming-Wei Chang, Kenton Lee, and Kristina Toutanova. 2019.
\newblock Bert: Pre-training of deep bidirectional transformers for language understanding.
\newblock In \emph{Proceedings of the 2019 conference of the North American chapter of the association for computational linguistics: human language technologies, volume 1 (long and short papers)}, pages 4171--4186.

\bibitem[{Emam et~al.(2021)Emam, Kondrich, Harrison, Lau, Wang, Kim, and Branson}]{emam2021state}
Zeyad Emam, Andrew Kondrich, Sasha Harrison, Felix Lau, Yushi Wang, Aerin Kim, and Elliot Branson. 2021.
\newblock On the state of data in computer vision: Human annotations remain indispensable for developing deep learning models.
\newblock \emph{arXiv preprint arXiv:2108.00114}.

\bibitem[{Flamholz et~al.(2024)Flamholz, Biller, and Kelly}]{flamholz2024large}
Zachary~N Flamholz, Steven~J Biller, and Libusha Kelly. 2024.
\newblock Large language models improve annotation of prokaryotic viral proteins.
\newblock \emph{Nature Microbiology}, 9(2):537--549.

\bibitem[{Glocker et~al.(2025)Glocker, H{\"o}nig, Hirschmanner, and Vincze}]{glocker2025llm}
Marc Glocker, Peter H{\"o}nig, Matthias Hirschmanner, and Markus Vincze. 2025.
\newblock Llm-empowered embodied agent for memory-augmented task planning in household robotics.
\newblock \emph{arXiv preprint arXiv:2504.21716}.

\bibitem[{Grattafiori et~al.(2024)Grattafiori, Dubey, Jauhri, Pandey, Kadian, Al-Dahle, Letman, Mathur, Schelten, Vaughan et~al.}]{grattafiori2024llama}
Aaron Grattafiori, Abhimanyu Dubey, Abhinav Jauhri, Abhinav Pandey, Abhishek Kadian, Ahmad Al-Dahle, Aiesha Letman, Akhil Mathur, Alan Schelten, Alex Vaughan, et~al. 2024.
\newblock The llama 3 herd of models.
\newblock \emph{arXiv preprint arXiv:2407.21783}.

\bibitem[{Guo et~al.(2025)Guo, Yang, Zhang, Song, Zhang, Xu, Zhu, Ma, Wang, Bi et~al.}]{guo2025deepseek}
Daya Guo, Dejian Yang, Haowei Zhang, Junxiao Song, Ruoyu Zhang, Runxin Xu, Qihao Zhu, Shirong Ma, Peiyi Wang, Xiao Bi, et~al. 2025.
\newblock Deepseek-r1: Incentivizing reasoning capability in llms via reinforcement learning.
\newblock \emph{arXiv preprint arXiv:2501.12948}.

\bibitem[{He et~al.(2024)He, Zannettou, Shen, and Zhang}]{he2024you}
Xinlei He, Savvas Zannettou, Yun Shen, and Yang Zhang. 2024.
\newblock You only prompt once: On the capabilities of prompt learning on large language models to tackle toxic content.
\newblock In \emph{2024 IEEE Symposium on Security and Privacy (SP)}, pages 770--787. IEEE.

\bibitem[{Jadhav et~al.(2024)Jadhav, Shanbhag, Thakurdesai, Sinare, and Joshi}]{jadhav2024limitations}
Suramya Jadhav, Abhay Shanbhag, Amogh Thakurdesai, Ridhima Sinare, and Raviraj Joshi. 2024.
\newblock On limitations of llm as annotator for low resource languages.
\newblock \emph{arXiv preprint arXiv:2411.17637}.

\bibitem[{Jiang et~al.(2024)Jiang, Wang, Shen, Kim, and Kim}]{jiang2024survey}
Juyong Jiang, Fan Wang, Jiasi Shen, Sungju Kim, and Sunghun Kim. 2024.
\newblock A survey on large language models for code generation.
\newblock \emph{arXiv preprint arXiv:2406.00515}.

\bibitem[{Jiang et~al.(2011)Jiang, Yu, Zhou, Liu, and Zhao}]{jiang2011target}
Long Jiang, Mo~Yu, Ming Zhou, Xiaohua Liu, and Tiejun Zhao. 2011.
\newblock Target-dependent twitter sentiment classification.
\newblock In \emph{Proceedings of the 49th annual meeting of the association for computational linguistics: human language technologies}, pages 151--160.

\bibitem[{Laskar et~al.(2023)Laskar, Rahman, Jahan, Hoque, and Huang}]{laskar2023can}
Md~Tahmid~Rahman Laskar, Mizanur Rahman, Israt Jahan, Enamul Hoque, and Jimmy Huang. 2023.
\newblock Can large language models fix data annotation errors? an empirical study using debatepedia for query-focused text summarization.
\newblock In \emph{Findings of the Association for Computational Linguistics: EMNLP 2023}, pages 10245--10255.

\bibitem[{Li et~al.(2024{\natexlab{a}})Li, Fan, Atreja, and Hemphill}]{li2024hot}
Lingyao Li, Lizhou Fan, Shubham Atreja, and Libby Hemphill. 2024{\natexlab{a}}.
\newblock “hot” chatgpt: The promise of chatgpt in detecting and discriminating hateful, offensive, and toxic comments on social media.
\newblock \emph{ACM Transactions on the Web}, 18(2):1--36.

\bibitem[{Li et~al.(2024{\natexlab{b}})Li, Liu, He, Zhang, Zhang, Ye, Lu, and Huang}]{li2024optimizing}
Rui Li, Qi~Liu, Liyang He, Zheng Zhang, Hao Zhang, Shengyu Ye, Junyu Lu, and Zhenya Huang. 2024{\natexlab{b}}.
\newblock Optimizing code retrieval: High-quality and scalable dataset annotation through large language models.
\newblock In \emph{Proceedings of the 2024 Conference on Empirical Methods in Natural Language Processing}, pages 2053--2065.

\bibitem[{Li et~al.(2025)Li, Wang, Sundaram, and Liu}]{li2025zero}
Yunzhe Li, Junting Wang, Hari Sundaram, and Zhining Liu. 2025.
\newblock A zero-shot generalization framework for llm-driven cross-domain sequential recommendation.
\newblock \emph{arXiv preprint arXiv:2501.19232}.

\bibitem[{Liu et~al.(2024)Liu, Feng, Xue, Wang, Wu, Lu, Zhao, Deng, Zhang, Ruan et~al.}]{liu2024deepseek}
Aixin Liu, Bei Feng, Bing Xue, Bingxuan Wang, Bochao Wu, Chengda Lu, Chenggang Zhao, Chengqi Deng, Chenyu Zhang, Chong Ruan, et~al. 2024.
\newblock Deepseek-v3 technical report.
\newblock \emph{arXiv preprint arXiv:2412.19437}.

\bibitem[{Liu et~al.(2025)Liu, Wu, Li, Shao, Pang, and Feng}]{liu2025automated}
Xiao Liu, Zirui Wu, Jiayi Li, Zhicheng Shao, Xun Pang, and Yansong Feng. 2025.
\newblock Automated annotation of evolving corpora for augmenting longitudinal network data: A framework integrating large language models and expert knowledge.
\newblock \emph{arXiv preprint arXiv:2503.01672}.

\bibitem[{Liu et~al.(2019)Liu, Ott, Goyal, Du, Joshi, Chen, Levy, Lewis, Zettlemoyer, and Stoyanov}]{liu2019roberta}
Yinhan Liu, Myle Ott, Naman Goyal, Jingfei Du, Mandar Joshi, Danqi Chen, Omer Levy, Mike Lewis, Luke Zettlemoyer, and Veselin Stoyanov. 2019.
\newblock Roberta: A robustly optimized bert pretraining approach.
\newblock \emph{arXiv preprint arXiv:1907.11692}.

\bibitem[{Mo et~al.(2024)Mo, Qin, Dong, Zhu, and Li}]{mo2024large}
Yuhong Mo, Hao Qin, Yushan Dong, Ziyi Zhu, and Zhenglin Li. 2024.
\newblock Large language model (llm) ai text generation detection based on transformer deep learning algorithm.
\newblock \emph{arXiv preprint arXiv:2405.06652}.

\bibitem[{Rasmussen et~al.(2022)Rasmussen, Kirk, and Moeslund}]{rasmussen2022challenge}
Christoffer~B{\o}gelund Rasmussen, Kristian Kirk, and Thomas~B Moeslund. 2022.
\newblock The challenge of data annotation in deep learning—a case study on whole plant corn silage.
\newblock \emph{Sensors}, 22(4):1596.

\bibitem[{Roziere et~al.(2023)Roziere, Gehring, Gloeckle, Sootla, Gat, Tan, Adi, Liu, Sauvestre, Remez et~al.}]{roziere2023code}
Baptiste Roziere, Jonas Gehring, Fabian Gloeckle, Sten Sootla, Itai Gat, Xiaoqing~Ellen Tan, Yossi Adi, Jingyu Liu, Romain Sauvestre, Tal Remez, et~al. 2023.
\newblock Code llama: Open foundation models for code.
\newblock \emph{arXiv preprint arXiv:2308.12950}.

\bibitem[{Sun et~al.(2024)Sun, Xu, Liu, Luan, Wang, Shang, Wen, and Yan}]{sun2024determlr}
Hongda Sun, Weikai Xu, Wei Liu, Jian Luan, Bin Wang, Shuo Shang, Ji-Rong Wen, and Rui Yan. 2024.
\newblock Determlr: Augmenting llm-based logical reasoning from indeterminacy to determinacy.
\newblock In \emph{Proceedings of the 62nd Annual Meeting of the Association for Computational Linguistics (Volume 1: Long Papers)}, pages 9828--9862.

\bibitem[{Taori et~al.(2023)Taori, Gulrajani, Zhang, Dubois, Li, Guestrin, Liang, and Hashimoto}]{alpaca}
Rohan Taori, Ishaan Gulrajani, Tianyi Zhang, Yann Dubois, Xuechen Li, Carlos Guestrin, Percy Liang, and Tatsunori~B. Hashimoto. 2023.
\newblock Stanford alpaca: An instruction-following llama model.
\newblock \url{https://github.com/tatsu-lab/stanford_alpaca}.

\bibitem[{Team et~al.(2025)Team, Du, Gao, Xing, Jiang, Chen, Li, Xiao, Du, Liao et~al.}]{team2025kimi}
Kimi Team, Angang Du, Bofei Gao, Bowei Xing, Changjiu Jiang, Cheng Chen, Cheng Li, Chenjun Xiao, Chenzhuang Du, Chonghua Liao, et~al. 2025.
\newblock Kimi k1. 5: Scaling reinforcement learning with llms.
\newblock \emph{arXiv preprint arXiv:2501.12599}.

\bibitem[{Tekumalla and Banda(2023)}]{tekumalla2023leveraging}
Ramya Tekumalla and Juan~M Banda. 2023.
\newblock Leveraging large language models and weak supervision for social media data annotation: an evaluation using covid-19 self-reported vaccination tweets.
\newblock In \emph{International Conference on Human-Computer Interaction}, pages 356--366. Springer.

\bibitem[{Touvron et~al.(2023)Touvron, Martin, Stone, Albert, Almahairi, Babaei, Bashlykov, Batra, Bhargava, Bhosale et~al.}]{touvron2023llama}
Hugo Touvron, Louis Martin, Kevin Stone, Peter Albert, Amjad Almahairi, Yasmine Babaei, Nikolay Bashlykov, Soumya Batra, Prajjwal Bhargava, Shruti Bhosale, et~al. 2023.
\newblock Llama 2: Open foundation and fine-tuned chat models.
\newblock \emph{arXiv preprint arXiv:2307.09288}.

\bibitem[{Van~Aken et~al.(2018)Van~Aken, Risch, Krestel, and L{\"o}ser}]{van2018challenges}
Betty Van~Aken, Julian Risch, Ralf Krestel, and Alexander L{\"o}ser. 2018.
\newblock Challenges for toxic comment classification: An in-depth error analysis.
\newblock \emph{arXiv preprint arXiv:1809.07572}.

\bibitem[{Wang and Chen(2023)}]{wang2023review}
Jianxun Wang and Yixiang Chen. 2023.
\newblock A review on code generation with llms: Application and evaluation.
\newblock In \emph{2023 IEEE International Conference on Medical Artificial Intelligence (MedAI)}, pages 284--289. IEEE.

\bibitem[{Wei et~al.(2022)Wei, Wang, Schuurmans, Bosma, Xia, Chi, Le, Zhou et~al.}]{wei2022chain}
Jason Wei, Xuezhi Wang, Dale Schuurmans, Maarten Bosma, Fei Xia, Ed~Chi, Quoc~V Le, Denny Zhou, et~al. 2022.
\newblock Chain-of-thought prompting elicits reasoning in large language models.
\newblock \emph{Advances in neural information processing systems}, 35:24824--24837.

\bibitem[{Wu et~al.(2024)Wu, Wang, and Jia}]{wu2024enhancing}
Jianfei Wu, Xubin Wang, and Weijia Jia. 2024.
\newblock Enhancing text annotation through rationale-driven collaborative few-shot prompting.
\newblock \emph{arXiv preprint arXiv:2409.09615}.

\bibitem[{Wu et~al.(2023)Wu, Fei, Qu, Ji, and Chua}]{wu2023next}
Shengqiong Wu, Hao Fei, Leigang Qu, Wei Ji, and Tat-Seng Chua. 2023.
\newblock Next-gpt: Any-to-any multimodal llm.
\newblock \emph{arXiv preprint arXiv:2309.05519}.

\bibitem[{Xu et~al.(2023)Xu, Xu, Wang, Liu, Zhu, and McAuley}]{xu2023small}
Canwen Xu, Yichong Xu, Shuohang Wang, Yang Liu, Chenguang Zhu, and Julian McAuley. 2023.
\newblock Small models are valuable plug-ins for large language models.
\newblock \emph{arXiv preprint arXiv:2305.08848}.

\bibitem[{Xu et~al.(2025)Xu, Shi, Zhuang, Yu, Ho, Wang, and Yang}]{xu2025collab}
Ran Xu, Wenqi Shi, Yuchen Zhuang, Yue Yu, Joyce~C Ho, Haoyu Wang, and Carl Yang. 2025.
\newblock Collab-rag: Boosting retrieval-augmented generation for complex question answering via white-box and black-box llm collaboration.
\newblock \emph{arXiv preprint arXiv:2504.04915}.

\bibitem[{Yadav et~al.(2024)Yadav, Choppa, and Schlechtweg}]{yadav2024towards}
Sachin Yadav, Tejaswi Choppa, and Dominik Schlechtweg. 2024.
\newblock Towards automating text annotation: A case study on semantic proximity annotation using gpt-4.
\newblock \emph{arXiv preprint arXiv:2407.04130}.

\bibitem[{Yang et~al.(2023)Yang, Xiao, Wang, Zhang, Bian, Yin, Lv, Pan, Wang, Yan et~al.}]{yang2023baichuan}
Aiyuan Yang, Bin Xiao, Bingning Wang, Borong Zhang, Ce~Bian, Chao Yin, Chenxu Lv, Da~Pan, Dian Wang, Dong Yan, et~al. 2023.
\newblock Baichuan 2: Open large-scale language models.
\newblock \emph{arXiv preprint arXiv:2309.10305}.

\bibitem[{Yang et~al.(2024)Yang, Yang, Zhang, Hui, Zheng, Yu, Li, Liu, Huang, Wei et~al.}]{yang2024qwen2}
An~Yang, Baosong Yang, Beichen Zhang, Binyuan Hui, Bo~Zheng, Bowen Yu, Chengyuan Li, Dayiheng Liu, Fei Huang, Haoran Wei, et~al. 2024.
\newblock Qwen2. 5 technical report.
\newblock \emph{arXiv preprint arXiv:2412.15115}.

\bibitem[{Ye et~al.(2025)Ye, Huang, Xiao, Chern, Xia, and Liu}]{ye2025limo}
Yixin Ye, Zhen Huang, Yang Xiao, Ethan Chern, Shijie Xia, and Pengfei Liu. 2025.
\newblock Limo: Less is more for reasoning.
\newblock \emph{arXiv preprint arXiv:2502.03387}.

\bibitem[{Zhang et~al.(2024)Zhang, Wang, Hua, Qi, Ding, and Zhou}]{zhang2024cogenesis}
Kaiyan Zhang, Jianyu Wang, Ermo Hua, Biqing Qi, Ning Ding, and Bowen Zhou. 2024.
\newblock Cogenesis: A framework collaborating large and small language models for secure context-aware instruction following.
\newblock \emph{arXiv preprint arXiv:2403.03129}.

\bibitem[{Zheng et~al.(2025)Zheng, Chen, Zhang, Kundu, Li, Liu, Xing, Wang, and Yao}]{zheng2025citer}
Wenhao Zheng, Yixiao Chen, Weitong Zhang, Souvik Kundu, Yun Li, Zhengzhong Liu, Eric~P Xing, Hongyi Wang, and Huaxiu Yao. 2025.
\newblock Citer: Collaborative inference for efficient large language model decoding with token-level routing.
\newblock \emph{arXiv preprint arXiv:2502.01976}.

\end{thebibliography}


\appendix

\clearpage
{\LARGE \bf Appendix}

\renewcommand\thesection{\Alph{section}}
\renewcommand\thefigure{\Alph{figure}}
\renewcommand\thetable{\Alph{table}}
\setcounter{section}{0}
\setcounter{figure}{0}
\setcounter{table}{0}

\section*{Organization of the Appendix}
The Appendix is organized as follows. 
\begin{itemize}
    \item Section~\ref{sec:Dataset_Description} introduces the dataset we used.
    \item Section~\ref{sec:appendix_prompt_details} provides the prompts we used.
    \item Section~\ref{sec:appendix_codes_details} provides the codes generated by LLMs.
\end{itemize}
    
\section{Dataset Description}
\label{sec:Dataset_Description}
Here, we introduce the two types of datasets used in this study. As for sentiment classification tasks, three public datasets are used, namely \textbf{mteb/tweet\_sentiment\_extraction}\footnote{https://huggingface.co/datasets/mteb/tweet\_sentiment\_extraction}, \textbf{jppgks/twitter-financial-news-sentiment}\footnote{https://huggingface.co/datasets/jppgks/twitter-financial-news-sentiment} and \textbf{LYTinn/sentiment-analysis-tweet}\footnote{https://huggingface.co/datasets/LYTinn/sentiment-analysis-tweet}. As for toxicity classification tasks, three public datasets are used, namely \textbf{karthikarunr/Cyberbullying-Toxicity-Tweets}\footnote{https://huggingface.co/datasets/karthikarunr/Cyberbullying-Toxicity-Tweets}, \textbf{jiaxin-wen/Implicit-Toxicity}\footnote{https://huggingface.co/datasets/jiaxin-wen/Implicit-Toxicity} and \textbf{heegyu/toxic\_conversations\_balanced}\footnote{https://huggingface.co/datasets/heegyu/toxic\_conversations\_balanced}.

\begin{itemize}
    \item \textbf{mteb/tweet\_sentiment\_extraction}: $24,739$ samples from its training set (train-$00000$-of-$00001$.parquet) are split into a $7$:$3$ ratio, yielding $17,317$ training and $7,422$ test samples.
    \item \textbf{jppgks/twitter-financial-news-sentiment}: The full training set ($9,543$ samples) and test set ($2,388$ samples) are directly adopted as the framework’s training and test sets.
    \item \textbf{LYTinn/sentiment-analysis-tweet}: The entire training set ($11,536$ samples) and test set ($3,377$ samples) are directly adopted as the framework’s training and test sets.
    \item \textbf{karthikarunr/Cyberbullying-Toxicity-Tweets}: $23,005$ samples from its training set (train-$00000$-of-$00001$.parquet) are split into $16,104$ training and $6,901$ test samples ($7$:$3$ ratio).
    \item \textbf{jiaxin-wen/Implicit-Toxicity}: $22,426$ samples from train/sft-train.json are divided into $15,698$ training and $6,728$ test samples ($7$:$3$ ratio).
    \item \textbf{heegyu/toxic\_conversations\_balanced}: $23,820$ samples from train.csv are split into $16,674$ training and $7,146$ test samples ($7$:$3$ ratio).
\end{itemize}

\section{Prompts}\label{sec:appendix_prompt_details}
We provide the prompts used in the paper in \Cref{table:sysprompt} and \Cref{table:apiprompt}.

\begin{table*}[h!]
  \centering
  \small
  \begin{adjustbox}{max width=\textwidth} 
  \begin{tabular}{l m{15cm}} 
    \toprule
    \multicolumn{2}{c}{System Prompt}  \\ 
    \midrule
    Model Selection &  \makecell[l]{Now I need you to help me write a code. Note that you need to strictly follow my requirements and the content \\you generate should be directly runnable code. \\
    Requirements:\\
    1. Model search: Find text annotation models similar to BERT on Hugging Face according to the text annotation \\requirements of sentiment classification (only supporting positive, negative, neutral and their respective confidence\\ scores) and toxic content detection (only supporting toxic and its confidence score). Note that models with multiple \\labels like unitary/toxic-bert do not meet the requirements. \\
    2. Quantity requirements: 3 sentiment classification models and 3 toxic content detection models.\\
    3. After selection, set up a UI interface for me to view the information of the selected models (using tkinter). \\
    4. Save the table in the UI interface to the local path \{config.path\}.\\
    Note:\\
    1. The table is for reference only. The number of models, model parameters, and HF downloads are inaccurate and \\need to be investigated by you.\\
    2. You need to search for models that meet the conditions and remember them. Your code only includes the UI part\\ (including models that meet the conditions).\\
    3. Please strictly follow the above requirements when writing the code.\\
    4. Your answer should only contain the code and nothing else.\\
    } \\
    \midrule
    Model Deployment & \makecell[l]{Now I need you to help me write a code. Note that you need to strictly follow my requirements and the content you\\ generate should be directly runnable code.
    I need to deploy a Hugging Face model locally. The model address is\\ https://hf-mirror.com/models. Please help me complete the following tasks:\\
    1. Model ID acquisition: Get the model IDs I want to download from the 'Model ID' column in the file \{Local File\}.\\
    2. Save path: I want to save the models to the local directory \{Local Path\}. For example, save \\'cardiffnlp\_twitter-roberta-base-sentiment-latest' to \{Local Path\}.\\
    3. Code generation: Please generate a complete and directly runnable Python code to download the models locally.\\
    Please generate the code according to the above requirements to complete the deployment.\\
    Note: Your answer should only contain the code and nothing else.} \\
    \midrule
    Data Annotation & \makecell[l]{Now I need you to help me write a code. Note that you need to strictly follow my requirements and the content\\ you generate should be directly runnable code. Your task is to help me call a pre-trained BERT model (or similar\\ models) from the local and use this model to label text data. The following are the specific requirements of the task:\\
    Task description:\\
    1. Model type: Use a pre-trained model like BERT or similar ones (e.g., RoBERTa, DistilBERT, etc.).\\
    2. Model ID: \{Model ID\}.\\
    3. Local model address: \{Local Path\}.\\
    4. Task type: \{Task Type\}.\\
    5. Input data: The address of the data file in the first column of the local xlsx file is \{Local File\}.\\
    6. Output result: The labeled result. If there is a confidence score, it is also required. Save it to \{Local File\}.\\
    7. Requirements for the result file: Change the column name of the label column in the saved file to \{label\_col\_name\},\\ and change the column name of the confidence score column to \{confidence\_col\_name\}'.
    } \\
    \midrule
    Model Fine-tuning  & \makecell[l]{Now I need you to help me write a code. Note that you must strictly follow my requirements, and the content you \\generate should be directly runnable code. Your task is to write code to perform full-parameter fine-tuning on the \\ \{Model ID\} model. The following are the specific requirements of the task:\\
    Task description:\\
    1. Model type: Use a pre-trained model like BERT or similar ones (e.g., RoBERTa, DistilBERT, etc.).\\
    2. Model ID: \{Model ID\}.\\
    3. Local model address: \{Local Path\}.\\
    4. Fine-tuning data: The 'text' column in the \{Local File\} is the text column of the fine-tuning data, and the 'label' \\column is the label column of the fine-tuning data.\\
    5. Save address for the fine-tuned model: \{Local Path\}.\\
    Note: Your answer should only contain the code and nothing else.} \\ 
    \bottomrule
  \end{tabular}
  \end{adjustbox}
  \caption{System Prompt for Data Annotation.}
  \label{table:sysprompt}
\end{table*}

\begin{table*}[h!]
  \centering
  \small
  \begin{adjustbox}{max width=\textwidth} 
  \begin{tabular}{l m{15cm}} 
    \toprule
    \multicolumn{2}{c}{Prompt}  \\ 
    \midrule
    Sentiment Classification &  \makecell[l]{You are an autoclassifier that's responsible for labeling input text. You must respond with only one of these labels: \\positive, negative, neutral. \\
    } \\
    \midrule
    Toxicity Classification &  \makecell[l]{You are an autoclassifier that's responsible for labeling input text. You must respond with only one of these labels: \\toxic, non-toxic.\\ } \\
    \bottomrule
  \end{tabular}
  \end{adjustbox}
  \caption{Prompt for Data Annotation.}
  \label{table:apiprompt}
\end{table*}

\section{Codes Generated by LLMs}\label{sec:appendix_codes_details}
We provide the codes generate by LLMs in \Cref{algo:model selection}, \Cref{algo:model deployment}, \Cref{algo:model labeling} and \Cref{algo:model finetuning}.

\begin{algorithm*}[h]
   \caption{PyTorch implementation of model selection generated by LLMs.}
   \label{algo:model selection}
    \definecolor{codeblue}{rgb}{0.25,0.5,0.5}
    \definecolor{codekw}{rgb}{0.85, 0.18, 0.50}
    \lstset{
      basicstyle=\fontsize{7.2pt}{7.2pt}\ttfamily,
      commentstyle=\fontsize{7.2pt}{7.2pt}\color{codeblue},
      keywordstyle=\fontsize{7.2pt}{7.2pt}\color{codekw},
    }
\begin{lstlisting}[language=python]
import tkinter as tk
from tkinter import ttk
import pandas as pd
import os

# Selected models based on the requirements
models = [
    {"Serial Number": 1, "Model ID": "cardiffnlp/twitter-roberta-base-sentiment-latest", 
    "Type": "sentiment", "Parameters": "125M", "HF Downloads": "2.43M"},
    {"Serial Number": 2, "Model ID": "cardiffnlp/twitter-xlm-roberta-base-sentiment", 
    "Type": "sentiment", "Parameters": "125M", "HF Downloads": "2.06M"},
    {"Serial Number": 3, "Model ID": "finiteautomata/bertweet-base-sentiment-analysis", 
    "Type": "sentiment", "Parameters": "110M", "HF Downloads": "1.06M"},
    {"Serial Number": 4, "Model ID": "s-nlp/roberta_toxicity_classifier", 
    "Type": "toxic", "Parameters": "110M", "HF Downloads": "160K"},
    {"Serial Number": 5, "Model ID": "JungleLee/bert-toxic-comment-classification", 
    "Type": "toxic", "Parameters": "110M", "HF Downloads": "46.3K"},
    {"Serial Number": 6, "Model ID": "garak-llm/toxic-comment-model", 
    "Type": "toxic", "Parameters": "110M", "HF Downloads": "9.19K"}
]

# Convert to DataFrame
df = pd.DataFrame(models)

# Function to save the table to Excel
def save_to_excel():
    save_path = 'C:\\Users\\21888\\Desktop\\1-24\\model.xlsx'
    df.to_excel(save_path, index=False)
    status_label.config(text=f"Table saved to {save_path}")

# Create the main window
root = tk.Tk()
root.title("Hugging Face Model Search")

# Create a frame for the table
frame = ttk.Frame(root)
frame.pack(pady=20)

# Create a treeview to display the table
tree = ttk.Treeview(frame, columns=list(df.columns), show="headings")
for col in df.columns:
    tree.heading(col, text=col)
    tree.column(col, width=100)
tree.pack()

# Insert data into the treeview
for index, row in df.iterrows():
    tree.insert("", "end", values=list(row))

# Create a button to save the table
save_button = ttk.Button(root, text="Save to Excel", command=save_to_excel)
save_button.pack(pady=10)

# Status label
status_label = ttk.Label(root, text="")
status_label.pack()

# Run the application
root.mainloop()
\end{lstlisting}
\end{algorithm*}

\begin{algorithm*}[h]
   \caption{PyTorch implementation of model deployment generated by LLMs.}
   \label{algo:model deployment}
    \definecolor{codeblue}{rgb}{0.25,0.5,0.5}
    \definecolor{codekw}{rgb}{0.85, 0.18, 0.50}
    \lstset{
      basicstyle=\fontsize{7.2pt}{7.2pt}\ttfamily,
      commentstyle=\fontsize{7.2pt}{7.2pt}\color{codeblue},
      keywordstyle=\fontsize{7.2pt}{7.2pt}\color{codekw},
    }
\begin{lstlisting}[language=python]
import pandas as pd
from transformers import AutoTokenizer, AutoModel
import os

def get_model_ids():
    file_path = r'C:\Users\21888\Desktop\1-24\model.xlsx'
    df = pd.read_excel(file_path)
    model_ids = df['Model ID'].tolist()
    return model_ids

def download_and_save_models(model_ids, save_base_dir):
    for model_id in model_ids:
        try:
            tokenizer = AutoTokenizer.from_pretrained(model_id)
            model = AutoModel.from_pretrained(model_id)

            model_save_dir = os.path.join(save_base_dir, model_id.replace("/", "_"))
            os.makedirs(model_save_dir, exist_ok=True)

            tokenizer.save_pretrained(model_save_dir)
            model.save_pretrained(model_save_dir)
            print(f"model {model_id} has been saved to {model_save_dir}")
        except Exception as e:
            print("error")

if __name__ == "__main__":
    model_ids = get_model_ids()

    save_base_dir = r'E:\model'

    download_and_save_models(model_ids, save_base_dir)
\end{lstlisting}
\end{algorithm*}

\begin{algorithm*}[h]
   \caption{PyTorch implementation of data annotation generated by LLMs.}
   \label{algo:model labeling}
    \definecolor{codeblue}{rgb}{0.25,0.5,0.5}
    \definecolor{codekw}{rgb}{0.85, 0.18, 0.50}
    \lstset{
      basicstyle=\fontsize{7.2pt}{7.2pt}\ttfamily,
      commentstyle=\fontsize{7.2pt}{7.2pt}\color{codeblue},
      keywordstyle=\fontsize{7.2pt}{7.2pt}\color{codekw},
    }
\begin{lstlisting}[language=python]
import torch
import pandas as pd
from transformers import AutoTokenizer, AutoModelForSequenceClassification

model_path = r'E:\model\cardiffnlp_twitter-roberta-base-sentiment-latest'
tokenizer = AutoTokenizer.from_pretrained(model_path)
model = AutoModelForSequenceClassification.from_pretrained(model_path)

input_file = r'C:\Users\21888\Desktop\test\11.xlsx'
df = pd.read_excel(input_file, header=None)

df = df.dropna(subset=[0])

texts = df[0].astype(str).tolist()

results = []

for text in texts:
    inputs = tokenizer(text, return_tensors='pt', truncation=True, padding=True)
    with torch.no_grad():
        logits = model(**inputs).logits
    probabilities = torch.softmax(logits, dim=-1)
    sentiment_label = torch.argmax(probabilities, dim=-1).item()
    sentiment_confidence = probabilities[0][sentiment_label].item()
    results.append({'sentiment_1': sentiment_label, 'sentiment_confidence_1': sentiment_confidence})

output_df = pd.DataFrame(results)
output_file = r'C:\Users\21888\Desktop\test\result_sentiment_1.xlsx'
output_df.to_excel(output_file, index=False)
\end{lstlisting}
\end{algorithm*}

\begin{algorithm*}[h]
   \caption{PyTorch implementation of model fine-tuning generated by LLMs.}
   \label{algo:model finetuning}
    \definecolor{codeblue}{rgb}{0.25,0.5,0.5}
    \definecolor{codekw}{rgb}{0.85, 0.18, 0.50}
    \lstset{
      basicstyle=\fontsize{7.2pt}{7.2pt}\ttfamily,
      commentstyle=\fontsize{7.2pt}{7.2pt}\color{codeblue},
      keywordstyle=\fontsize{7.2pt}{7.2pt}\color{codekw},
    }
\begin{lstlisting}[language=python]
import torch
from transformers import RobertaForSequenceClassification, RobertaTokenizer, Trainer, TrainingArguments
from sklearn.model_selection import train_test_split
import pandas as pd
from datasets import Dataset

data = pd.read_excel('C:\\Users\\21888\\Desktop\\test\\finetune_sentiment.xlsx')
texts = data['text'].tolist()
labels = data['label'].tolist()

train_texts, val_texts, train_labels, val_labels = train_test_split(texts, labels, test_size=0.2, random_state=42)

train_dataset = Dataset.from_dict({'text': train_texts, 'label': train_labels})
val_dataset = Dataset.from_dict({'text': val_texts, 'label': val_labels})

model_name = 'E:\\model\\cardiffnlp_twitter-roberta-base-sentiment-latest'
model = RobertaForSequenceClassification.from_pretrained(model_name)
tokenizer = RobertaTokenizer.from_pretrained(model_name)

def preprocess_function(examples):
    return tokenizer(examples['text'], truncation=True, padding='max_length', max_length=128)

train_dataset = train_dataset.map(preprocess_function, batched=True)
val_dataset = val_dataset.map(preprocess_function, batched=True)

training_args = TrainingArguments(
    output_dir='E:\\fine-tune-model\\cardiffnlp_twitter-roberta-base-sentiment-latest',
    evaluation_strategy="epoch",
    learning_rate=2e-5,
    per_device_train_batch_size=16,
    per_device_eval_batch_size=16,
    num_train_epochs=3,
    weight_decay=0.01,
    save_total_limit=2,
    save_steps=500,
    logging_dir='./logs',
    logging_steps=10,
)

trainer = Trainer(
    model=model,
    args=training_args,
    train_dataset=train_dataset,
    eval_dataset=val_dataset,
)

trainer.train()

trainer.save_model('E:\\fine-tune-model\\cardiffnlp_twitter-roberta-base-sentiment-latest')

tokenizer.save_pretrained('E:\\fine-tune-model\\cardiffnlp_twitter-roberta-base-sentiment-latest')
\end{lstlisting}
\end{algorithm*}

\end{document}